%% file: MAStyleTransfer.tex
\documentclass[sigconf]{acmart}

\newcommand{\final}{1}

\AtBeginDocument{%
  \providecommand\BibTeX{{%
    \normalfont B\kern-0.5em{\scshape i\kern-0.25em b}\kern-0.8em\TeX}}}

\usepackage{booktabs}
\usepackage{wasysym}
\usepackage{multirow}
\usepackage{dsfont}
\usepackage{bbm}
\usepackage{array,color}
\usepackage{subfigure}
\usepackage{verbatim}

\input{macros.tex}

\copyrightyear{2020}
\acmYear{2020}
\setcopyright{acmcopyright}\acmConference[MM '20]{Proceedings of the 28th ACM International Conference on Multimedia}{October 12--16, 2020}{Seattle, WA, USA}
\acmBooktitle{Proceedings of the 28th ACM International Conference on Multimedia (MM '20), October 12--16, 2020, Seattle, WA, USA}
\acmPrice{15.00}
\acmDOI{10.1145/3394171.3414015}
\acmISBN{978-1-4503-7988-5/20/10}

\begin{document}
\fancyhead{}

\title{Arbitrary Style Transfer via Multi-Adaptation Network}

\author{Yingying Deng}
\affiliation{%
  \institution{School of Artificial Intelligence, UCAS \& NLPR, Institute of Automation, CAS}
}
  \email{dengyingying2017@ia.ac.cn}

\author{Fan Tang}
\authornote{Co-corresponding authors}
\affiliation{%
  \institution{NLPR, Institute of Automation, CAS}
}
\email{tfan.108@gmail.com}

\author{Weiming Dong}
\authornotemark[1]
\affiliation{%
  \institution{NLPR, Institute of Automation, CAS \& CASIA-LLVision Joint Lab}
}
\email{weiming.dong@ia.ac.cn}

\author{Wen Sun}
\affiliation{%
 \institution{Institute of Automation, CAS \& School of Artificial Intelligence, UCAS}
 }
 \email{sunwen2017@ia.ac.cn}

\author{Feiyue Huang}
\affiliation{%
  \institution{Youtu Lab, Tencent}
  }
  \email{garyhuang@tencent.com}

\author{Changsheng Xu}
\affiliation{%
  \institution{NLPR, Institute of Automation, CAS \& CASIA-LLVision Joint Lab}
}
\email{csxu@nlpr.ia.ac.cn}

\renewcommand{\shortauthors}{Deng, et al.}

\input{abstract}


\maketitle
\input{Introduction}

\input{Relatedwork}

\input{Methodology}

\input{Experiment}

\input{Conclusion}

\begin{acks}
This work was supported by National Natural Science Foundation of China under nos. 61832016 and 61672520, and by CASIA-Tencent
Youtu joint research project.
\end{acks}

\bibliographystyle{ACM-Reference-Format}
\bibliography{MAStyleTransfer}

\end{document}

%% file: macros.tex
\usepackage{ifthen}
\definecolor{WeimingColor}{rgb}{0,0,0.8} 
\definecolor{XingjiaColor}{rgb}{0.0,0.8,0.8} 
\definecolor{FanColor}{rgb}{0.8,0,0.8}
\definecolor{OliverColor}{rgb}{0.6,0,0.0}

\newcommand{\weiming}[1]{{\color{WeimingColor} [Weiming: #1]}}

\newcommand{\xingjia}[1]{{\color{XingjiaColor}[Xingjia: #1]}}
\newcommand{\fan}[1]{{\color{FanColor}[Fan: #1]}}

\newcommand{\warning}[1]{{\it\color{red} #1}}
\newcommand{\toremove}[1]{{\it\color{red} (To remove) #1}}
\newcommand{\note}[1]{{\it\color{blue} #1}}
\newcommand{\nothing}[1]{}
\usepackage{engord}
\ifthenelse{\equal{\final}{1}}
{
\renewcommand{\weiming}[1]{}
\renewcommand{\fan}[1]{}
\renewcommand{\xingjia}[1]{}

\renewcommand{\warning}[1]{}
\renewcommand{\toremove}[1]{}
\renewcommand{\note}[1]{}
\renewcommand{\nothing}[1]{}
}
{}

%% file: abstract.tex
\begin{abstract}
 Arbitrary style transfer is a significant topic with research value and application prospect. 
 A desired style transfer, given a content image and referenced style painting, would render the content image with the color tone and vivid stroke patterns of the style painting while synchronously maintaining the detailed content structure information. 
 Style transfer approaches would initially learn content and style representations of the content and style references and then generate the stylized images guided by these representations. 
 In this paper, we propose the multi-adaptation network which involves two self-adaptation (SA) modules and one co-adaptation (CA) module:
 the SA modules adaptively disentangle the content and style representations, i.e., content SA module uses position-wise self-attention to enhance content representation and style SA module uses channel-wise self-attention to enhance style representation;
 the CA module rearranges the distribution of style representation based on content representation distribution by calculating the local similarity between the disentangled content and style features in a non-local fashion.
 Moreover, a new disentanglement loss function enables our network to extract main style patterns and exact content structures to adapt to various input images, respectively.
 Various qualitative and quantitative experiments demonstrate that the proposed multi-adaptation network leads to better results than the state-of-the-art style transfer methods.
\end{abstract}

%% file: Introduction.tex
\section{Introduction}
Artistic style transfer is a significant technique that focuses on rendering natural images with artistic style patterns and maintaining the content structure of natural images at the same time. 
In recent years, researchers have applied convolutional neural networks (CNNs) to perform image translation and stylization~\citep{zhu:2017:unpaired,gatys:2016:image}. 
Gatys et al.~\citep{gatys:2016:image} innovatively used deep features extracted from VGG16 to represent image content structure and calculate the correlation of activation maps to obtain image style patterns. 
However, this optimization-based method is also time-consuming.
Based on \citep{gatys:2016:image}, many works either speed up the transfer procedure or promote the generation quality~\citep{gatys:2016:image,johnson:2016:perceptual,ulyanov:2016:texture,kolkin:2019:style,shen:2018:neural,zhi:2018:structure,wu:2018:direction}. 
Johnson et al.~\cite{johnson:2016:perceptual} used feed-forward neural networks to achieve the purpose of real-time style rendering. 
Gatys et al.~\citep{gatys:2017:controlling} improved the basic model of \citep{gatys:2016:image} to obtain higher-quality results and broaden the applications. 

To expand the application of style transfer further, many works focus on arbitrary style transfer methods~\cite{li:2017:universal,jing:2020:dynamic,wang:2017:zm,park:2019:arbitrary,li:2019:learning,yao:2019:attention,gu:2018:arbitrary,sheng:2018:avatar}.
AdaIN~\citep{Huang:2017:Arbitrary} and WCT ~\citep{li:2017:universal} align the second-order statistics of style image to content image.
However, the holistic style transfer process makes the generated quality disappointing.
Patch-swap based methods~\citep{chen:2016:swap, yao:2019:attention} aim to transfer style image patches to content image according to the similarity between patches pairs. However, when the distributions of content and style structure vary greatly, few style patterns are transferred to the content image through style-swap~\citep{chen:2016:swap}. 
Yao et al.~\citep{ yao:2019:attention} improved the style-swap~\citep{chen:2016:swap} method by adding multi-stroke control and self-attention mechanism. 
Inspired by self-attention mechanism, Park et al.~\cite{park:2019:arbitrary} proposed style-attention to match the style features onto the content features, but may encounter semantic structures of content image distorting problem.
Moreover, most style transfer methods use a common encoder to extract features of content and style images, which neglect the domain-specific features contributing to improved generation.

\begin{figure}[h]
\setlength{\abovecaptionskip}{2mm}
\setlength{\belowcaptionskip}{-2mm}
\centering
\includegraphics[width=\linewidth]{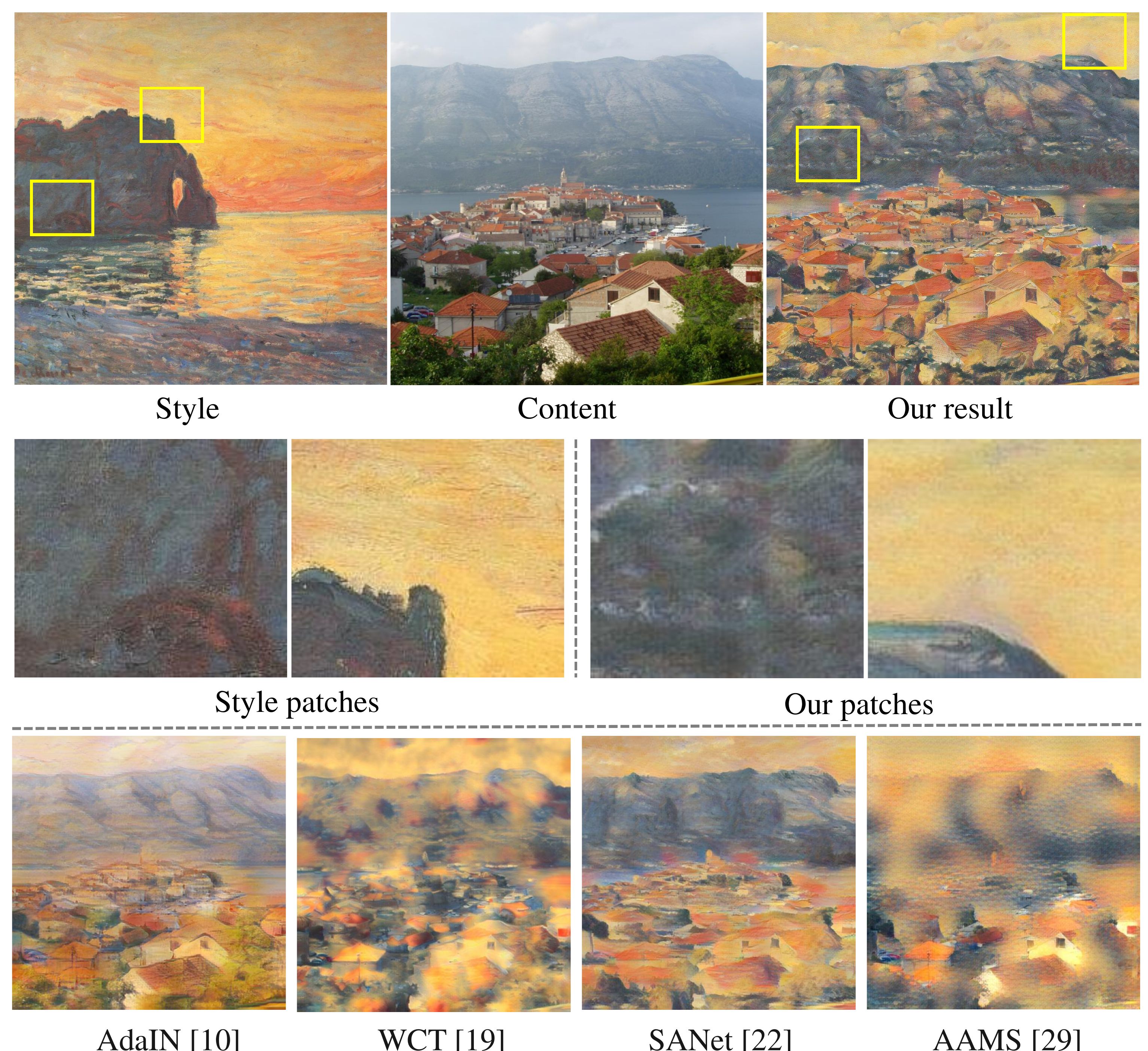}
\caption{Stylized result using Claude Monet's painting as style reference.
Compared with some state-of-the-art algorithms, our result can preserve detailed content structures and maintain vivid style patterns. For example, the mountain and house structures in our result are well preserved. We can observe that the painting strokes and color theme are well transferred by zooming in on similar patches in the style reference and our result.}
\vspace{-4mm}
\label{fig:p1}
\end{figure}

In recent years, some researchers~\citep{zhu:2017:unpaired,liu:2017:unsupervised,zhu:2017:toward,gonzalez-garcia:2018:image,kotovenko:2019:a,kotovenko:2019:content,yu:2019:multi,huang:2018:multimodal,kazemi:2019:style,Wang:2020:VCC} have used Generative Adversarial Networks (GANs) for high-quality image-to-image translation. GAN-based methods can generate high-quality artistic works that can be considered as real. 
The style and content representation are essential for the translation model. 
Numerous works~\citep{kotovenko:2019:content, zhu:2017:toward,gonzalez-garcia:2018:image,yu:2019:multi} focused on the disentanglement of style and content for allowing models to be aware of isolated factors of deep features. However, the image-to-image translation is difficult to adapt to arbitrary style transfer because of its limitation in unseen domain although it can achieve multi-modal style-guided translation results.

To enhance the generation effect of arbitrary style transfer methods aforementioned, we propose a flexible and efficient arbitrary style transfer model with disentanglement mechanism for preserving the detailed structures of the content images, while transferring rich style patterns of referenced paintings to generated results. 
As shown in Figure~\ref{fig:p1}, state-of-the-art methods can render the referenced color tone and style into the content image.
However, the content structures (outline of the houses) and stroke patterns are not well preserved and transferred.
In this work, we propose multi-adaptation network that involves two \textit{Self-Adaptation} (SA) modules and one \textit{Co-Adaptation} (CA) module. 
The SA module uses the position-wise self-attention to enhance content representation and channel-wise self-attention to enhance style representation, which adaptively disentangle the content and style representation.
Meanwhile, the CA module adjusts the style distribution to adapt to content distribution. 
Our model can learn effective content and style features through the interaction between SA and CA and rearrange the style features based on content features. 
Then, we merge the rearranged style and content features to obtain the generated results. 
Our method considers the global information in content and style image and the local similarity between image patches through the SA and CA procedure.
Moreover, we introduce a novel disentanglement loss for style and content representation.
The content disentanglement loss makes the content features extracted from stylized results similar when generating a series of stylized results by using a common content image and different style images. The style disentanglement loss makes the style features extracted from stylized results similar when generating a series of stylized results by using a common style image and different content images.
The disentanglement loss allows the network to extract main style patterns and exact content features to adapt to various content and style images, respectively.
In summary, our main contributions are as follows:
\begin{itemize}
\item  A flexible and efficient multi-adaptation arbitrary style transfer model involving two SA modules and one CA module.
\item  A novel disentanglement loss function for style and content disentanglement to extract well-directed style and content information.
\item  Various experiments illustrate that our method can preserve the detailed structures of the content images, and transfer rich style patterns of reference paintings to the generated results. 
Furthermore, we analyze the influence of different convolutional receptive field sizes on the CA module when calculating the local similarity between disentangled content and style features. 
\end{itemize}

%% file: Relatedwork.tex
\section{RELATED WORK}
\begin{figure*}[!tph]
\setlength{\abovecaptionskip}{2mm}
\centering
\includegraphics[width=\linewidth]{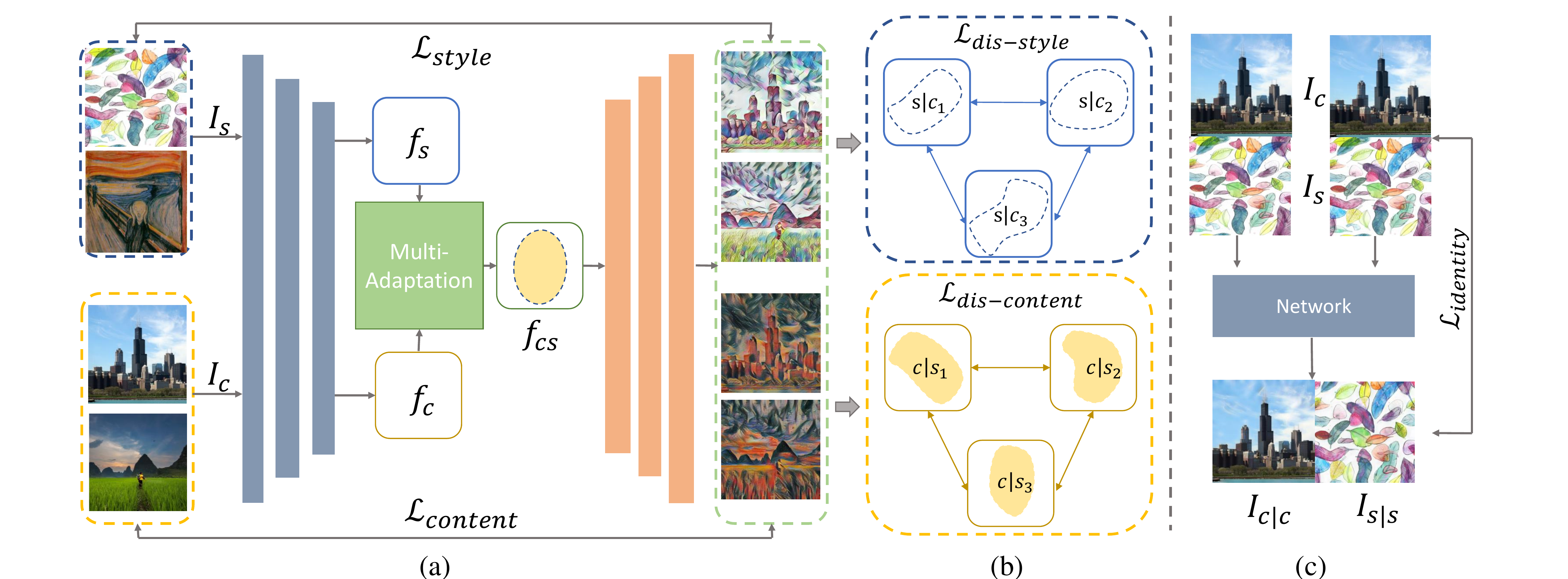}
\caption{(a) Structure of our network. The blue blocks are the encoder, and the orange blocks are the decoder. Given input content images $I_c$ and style images $I_s$, we can obtain corresponding features $f_c$ and $f_s$, respectively, through the encoder. Then we feed $f_c$ and $f_s$ to the multi-adaptation module and acquire the generated features $f_{cs}$. Finally, we generate the results $I_{cs}$ through the decoder. Moreover, the losses are calculated through a pretrained VGG19.
$\mathcal{L}_{content}$ measures the difference between $I_{cs}$ and $I_c$.
$\mathcal{L}_{style}$ calculates the difference between $I_{cs}$ and $I_s$.
(b) Disentanglement loss.     
 $\mathcal{L}_{dis-content}$ determines the content difference among stylized results, which are generated using different style images and a common content image.
 $\mathcal{L}_{dis-style}$ evaluates the style difference among stylized results, which are generated using different content images and a common style image. (c) Identity loss. $\mathcal{L}_{identify}$ quantifies the difference between $I_{c|c}(I_{s|s})$ and $I_c(I_s)$, where $I_{c|c}(I_{s|s})$ is the stylized result by using two common content (style) images.}
\vspace{-2mm}
\label{fig:network}
\end{figure*}

\paragraph{Style Transfer}
Since Gatys et al.~\citep{gatys:2016:image} proposed the first style transfer method by using CNNs, many works are devoted to promoting the transfer efficiency and generation effects. 
Some works~\citep{johnson:2016:perceptual,li:2016:precomputed,ulyanov:2016:texture} proposed real-time feed-forward style transfer networks, which can only transfer one kind of style by training an independent network.
Arbitrary style transfer becomes a major research topic to obtain a wide range of applications.
Chen et al.~\citep{chen:2016:swap} initially swapped the style image patch onto content images based on patch similarity and achieved fast style transfer for arbitrary style images.
Huang et al.~\citep{Huang:2017:Arbitrary} proposed adaptive instance normalization to adjust the mean and variance of content images to style images in a holistic fashion.
Li et al.~\citep{li:2017:universal} aligned the covariance of style and content images by using WCT and transferred multilevel style patterns to content images to obtain better-stylized results.
Avatar-Net~\citep{sheng:2018:avatar} applied style decorator to guarantee semantically aligned and holistically matching for combining the local and global style patterns to stylized results.
Park et al.~\citep{park:2019:arbitrary} proposed a style-attention network to match the style features onto the content features for achieving good results with evident style patterns.
Yao et al.~\citep{yao:2019:attention} achieved multi-stroke style results using self-attention mechanism.

However, the above arbitrary style transfer methods cannot efficiently balance content structure preservation and style pattern rendering. The disadvantages of these methods can be observed in Section~\ref{sec:CPW}.
Therefore, we aim to propose an arbitrary style transfer network that effectively transfers style patterns to content image while maintaining detailed content structures.
\paragraph{Feature Disentanglement}
In recent years, researches~\citep{zhu:2017:unpaired,liu:2017:unsupervised,zhu:2017:toward,gonzalez-garcia:2018:image,kotovenko:2019:a,kotovenko:2019:content,yu:2019:multi,huang:2018:multimodal,kazemi:2019:style} used generative adversarial networks (GANs), which can be applied to style transfer task in some cases, to achieve image-to-image translation.
A significant thought suitable for the style transfer task is that the style and content feature should be disentangled because of the domain deviation. 
Zhu et al.~\citep{huang:2018:multimodal} used two encoders to extract latent code to disentangle style and content representation. 
Kotovenko et al.~\cite{kotovenko:2019:content} proposed a disentanglement loss to separate style and content.
Kazemi et al.~\citep{kazemi:2019:style}  described a style and content disentangled GAN (SC-GAN) to learn a semantic representation of content and textual patterns of style. Yu et al.~\citep{yu:2019:multi} disentangled the input to latent code through an encoder-decoder network.

The existing disentanglement network usually adopted different encoders to decouple features through a training procedure. However, the structures of encoders are similar, which are not suitable enough for the disentanglement of style and content. In this paper, we design content and style SA modules to disentangle features specifically by considering the structures of content and texture of style.

%% file: Methodology.tex
\section{Methodology}
\label{sec:method}
\begin{figure*}[!tph]
\setlength{\abovecaptionskip}{2mm}
\centering
\includegraphics[width=\linewidth]{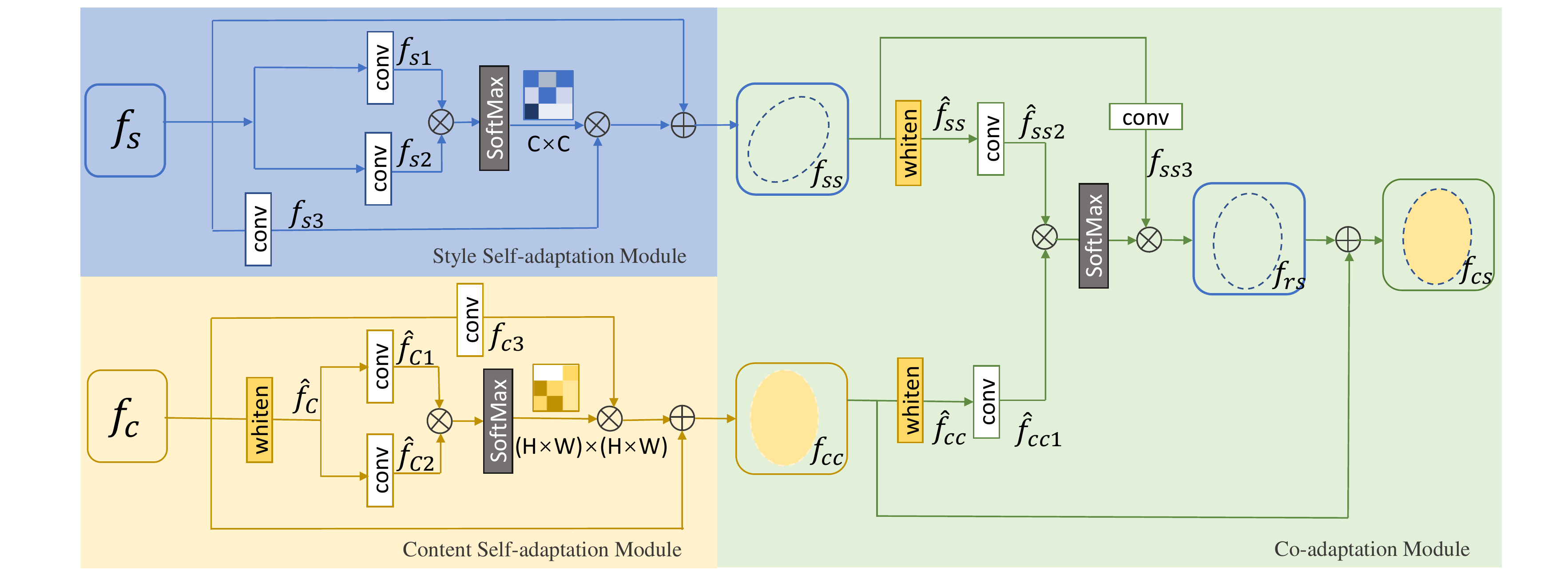}
\caption{Multi-adaptation Network. We disentangle the content and style through two independent SA modules.
We can disentangle the corresponding content/style representation $f_c(f_s)$ to $f_{cc}(f_{ss})$ through the content/style SA module. Then, the CA module rearranges the style distribution based on content distribution and generates rearranged features $f_{rs}$. Finally, we merge $f_{cc}$ and $f_{rs}$ to obtain stylized features $f_{cs}$.}
\vspace{-2mm}
\label{fig:SACA}
\end{figure*}
For the purpose of arbitrary style transfer, we propose a feed-forward network, which contains an encoder-decoder architecture and a multi-adaptation module.
Figure~\ref{fig:network} shows the structure of our network.
We use a pretrained VGG19 network as an encoder to extract deep features.
Given a content image $I_{c}$ and style image $I_{s}$, we can extract corresponding feature maps $f_c^i = \mathcal{E}(I_{c})$ and $f_s^i = \mathcal{E}(I_{s})$, $i \in \{1,...,L\}$.
However, the encoder is pretrained using ImageNet dataset for classification tasks, which is not suitable enough for style transfer tasks.
Meanwhile, only a few domain-specific features can be extracted using a common encoder given the domain deviation between artistic paintings and photographic images. 
Therefore, we proposed a multi-adaptation module to disentangle style and content representation through the self-adaptation process, and then rearrange the disentangled style distribution according to content distribution through the co-adaptation process. We can obtain stylized features $f_{cs}$ through the multi-adaptation module.
Section~\ref{sec:SACA} describes the multi-adaptation module in detail.
The decoder is a mirrored version of the encoder, and we can obtain the generated result $I_{cs} = \mathcal{D}(f_{cs})$.
The model is trained by minimizing three types of loss functions described in Section~\ref{sec:Loss}.

 
\subsection{Multi-adaptation Module}
\label{sec:SACA}
Figure~\ref{fig:SACA} shows the multi-adaptation module, which is divided into three parts: position-wise content SA module, channel-wise style SA module, and CA module.
We disentangle the content and style through two independent position-wise content and channel-wise style SA modules.
We can disentangle the corresponding content/style representation $f_c(f_s)$ to $f_{cc}(f_{ss})$ through the content/style SA module. 
Then the CA module rearranges the style representation based on content representation and generates stylized features $f_{cs}$.

\paragraph{Position-wise Content Self-adaptation Module}
Preserving the semantic structure of the content image in the stylized result is important, so we introduce position attention module in \citep{fu:2019:dual} to capture long-range information adaptively in content features. 
Given a content feature map $f_c \in \mathbb{R}^{C \times H \times W}$, $\hat{f_c}$ denotes the whitened content feature map, which removes textural information related to style by using whitening transform in ~\citep{li:2017:universal}. 
We feed $\hat{f_c}$ to two convolution layers, and generate two new feature maps $\hat{f}_{c1}$ and $\hat{f}_{c2}$. Meanwhile, we feed $f_c$ to another convolution layer to generate new feature map $f_{c3}$.
We reshape  $\hat{f}_{c1}$, $\hat{f}_{c2}$ and $f_{c3}$ to $ \mathbb{R}^{C \times N}$, where $N=H\times W$. Then, the content spatial attention map $A_c \in \mathbb{R}^{N \times N}$ is formulated as follows:
\begin{equation}
A_c= SoftMax({\hat{f}_{c1}}^T \otimes \hat{f}_{c2}),
\end{equation}
where $\otimes$ represents the matrix multiplication; and ${\hat{f}_{c1}}^T \otimes \hat{f}_{c2}$ denotes the position-wise multiplication between feature maps $\hat{f}_{c1}$ and $\hat{f}_{c2}$ .
Then, we obtain the enhanced content feature map $f_{c}$ through a matrix multiplication and element-wise addition:
\begin{equation}
f_{cc} = f_{c3}\otimes A_c^T + f_c.
\end{equation}

\paragraph{Channel-wise Style Self-adaptation Module}
Learning style patterns (e.g., texture and strokes) of the style image is important for style transfer.
Inspired by \citep{gatys:2016:image}, the channel-wise inner product between the vectorized feature maps can represent style, so we introduce channel attention module in \citep{fu:2019:dual} to enhance the style patterns in style images. The input style features do not need to be whitened, which is different from the content SA module.
We feed the style feature map $f_s \in \mathbb{R}^{C \times H \times W}$ to two convolution layers and generate two new feature maps $f_{s1}$ and $f_{s2}$. Meanwhile, we feed $f_s$ to another convolution layer to generate new feature map $f_{s3}$.
We reshape $f_{s1}$, $f_{s2}$ and $f_{s3}$ to $ \mathbb{R}^{C \times N}$, where $N=H\times W$. Then, the style spatial attention map $A_s \in \mathbb{R}^{C \times C}$ is formulated as follows:
\begin{equation}
A_s= SoftMax(f_{s1}  \otimes  f_{s2}^T),
\end{equation}
where $f_{s1} \otimes f_{s2}^T$ represents the channel-wise multiplication between feature maps $f_{s1}$ and $f_{s2}$.
Then, we adjust the style feature map $f_{s}$ through a matrix multiplication and an element-wise addition:
\begin{equation}
f_{ss} = A_s^T  \otimes  f_{s3} + f_s.
\end{equation}

\paragraph{Co-adaptation Module} 
Through the SA module, we obtain the disentangled style and content features. Then, we propose the CA module to calculate the correlation between the disentangled features, and recombine them adaptively onto an output feature map.
The generated results can not only retain the prominent content structure, but also adjust semantic content with the appropriate style patterns based on the correlation.
Figure~\ref{fig:SACA} shows the CA process.
Initially, the disentangled style feature map $f_{ss} $ and content feature map $f_{cc} $ are whitened to $\hat{f}_{ss}$ and $\hat{f}_{cc}$, respectively. 
Then, we feed $\hat{f}_{cc}$ and $\hat{f}_{ss}$ to two convolution layers to generate two new feature maps $\hat{f}_{cc1}$ and $\hat{f}_{ss2}$. 
Meanwhile, we feed feature map $f_{ss}$ to another convolution layer to generate a new feature map $f_{ss3}$.
We reshape $\hat{f}_{cc1}$, $\hat{f}_{ss2}$, and $f_{ss3}$ to $ \mathbb{R}^{C \times N}$, where $N=H\times W$. Then, the correlation map $A_{cs} \in \mathbb{R}^{N \times N}$ is formulated as follows:
\begin{equation}
A_{cs}= SoftMax({\hat{f}_{cc1}}^T \otimes \hat{f}_{ss2}),
\end{equation}
where the value of $A_{cs}$ in position $(i,j)$ measures the correlation between the $i$-th and the $j$-th position in content and style features, respectively.
Then, the rearranged style feature map $f_{rs}$ is mapped by:
\begin{equation}
f_{rs} = f_{ss3} \otimes A_{cs}^T.
\end{equation}
Finally, the CA result is achieved by:
\begin{equation}
f_{cs} = f_{rs} + f_{cc}.
\end{equation}

\subsection{Loss Function }
\label{sec:Loss}

Our network contains three loss functions in training procedure.

\paragraph{Perceptual Loss} Similar to AdaIN~\citep{Huang:2017:Arbitrary}, we use a pretrained VGG19 to compute the content and style perceptual loss.
The content perceptual loss $\mathcal{L}_{content}$ is used to minimize the content difference between generated and content images, where
\begin{equation}
\mathcal{L}^i_{content} = \lVert \phi_i(I_{cs}) - \phi_i(I_{c})  \lVert_2.
\end{equation}
The style perceptual loss $\mathcal{L}_{style}$ is utilized to minimize the style difference between generated and style images:
\begin{equation}
\mathcal{L}^i_{style} = \lVert \mu(\phi_i(I_{cs})) - \mu(\phi_i(I_{c}))  \lVert_2 
+  \lVert \sigma(\phi_i(I_{cs})) - \sigma(\phi_i(I_{c}))  \lVert_2,
\end{equation}
where $\phi_i(\cdot)$ denotes the features extracted from $i$-th layer in a pretrained VGG19, $\mu(\cdot)$ denotes the mean of features, and $\sigma (\cdot)$ is the variance of features.

\paragraph{Identity Loss}
Inspired by ~\citep{park:2019:arbitrary}, we introduce the identity loss to provide a soft constraint on the mapping relation between style and content features.
The identify loss is formulated as follows:
\begin{equation}
\mathcal{L}_{identity} = \lVert I_{c|c} - I_{c}  \lVert_2 + \lVert I_{s|s} - I_{s}  \lVert_2,
\end{equation}
where $I_{c|c}$ indicates the generated results using one natural image as content and style images simultaneously and $I_{s|s}$ signifies generated results using one painting as content and style images simultaneously.

\paragraph{Disentanglement Loss}
The style features should be independent from the target content to separate the style and content representation. 
That is, the content disentanglement loss renders the content features extracted from stylized results similar when generating a series of stylized results by using a common content image and different style images. 
The style disentanglement loss renders the style features extracted from stylized results similar when generating a series of stylized results by using a common style image and different content images. Therefore, we propose a novel disentanglement loss as follows:
\begin{equation}
\begin{split}
\mathcal{L}^i_{dis\_content} = \lVert \phi_i(I_{c|s_1}) - \phi_i(I_{c|s_2})  \lVert_2, \\
\mathcal{L}^i_{dis\_style} = \lVert \mu(\phi_i(I_{s|c_1})) - \mu(\phi_i(I_{s|c_2}))  \lVert_2 \\ 
+  \lVert \sigma(\phi_i(I_{s|c_1})) - \sigma(\phi_i(I_{s|c_1}))  \lVert_2,
\end{split}
\end{equation}
where $I_{c|s_1}$ and $I_{c|s_2}$ are the generated results by using a common content image and different style images, and $I_{s|c_1}$ and $I_{s|c_2}$ represent the generated results by using a common style image and different content images.
The total loss function is formulated as follows:
\begin{equation}
\begin{split}
\mathcal{L} =  \lambda_{c}\mathcal{L}^j_{content} +\lambda_{dis\_c}\mathcal{L}^j_{dis\_content}  + \lambda_{id}\mathcal{L}_{identity} \\
+ \lambda_{s}  \sum_{i=1}^{L} \mathcal{L}^i_{style}+ \lambda_{dis\_s}  \sum_{i=1}^{L} \mathcal{L}^i_{dis\_style}.
\end{split}
\end{equation}

In general, the loss functions constrain the global similarity between generated results and content/style images. 
The two SA modules calculate the long-range self-similarity of input features to  disentangle the global content/style representation.
The CA module rearranges the style distribution according to content distribution by calculating the local similarity between the disentangled content and style features in a non-local fashion.
Therefore, our network can consider the global content structure and local style patterns to generate fascinating results.

%% file: Experiment.tex
\section{Experiments}
\label{sec:experiments}


\begin{figure*}
\centering
\includegraphics[width=\linewidth]{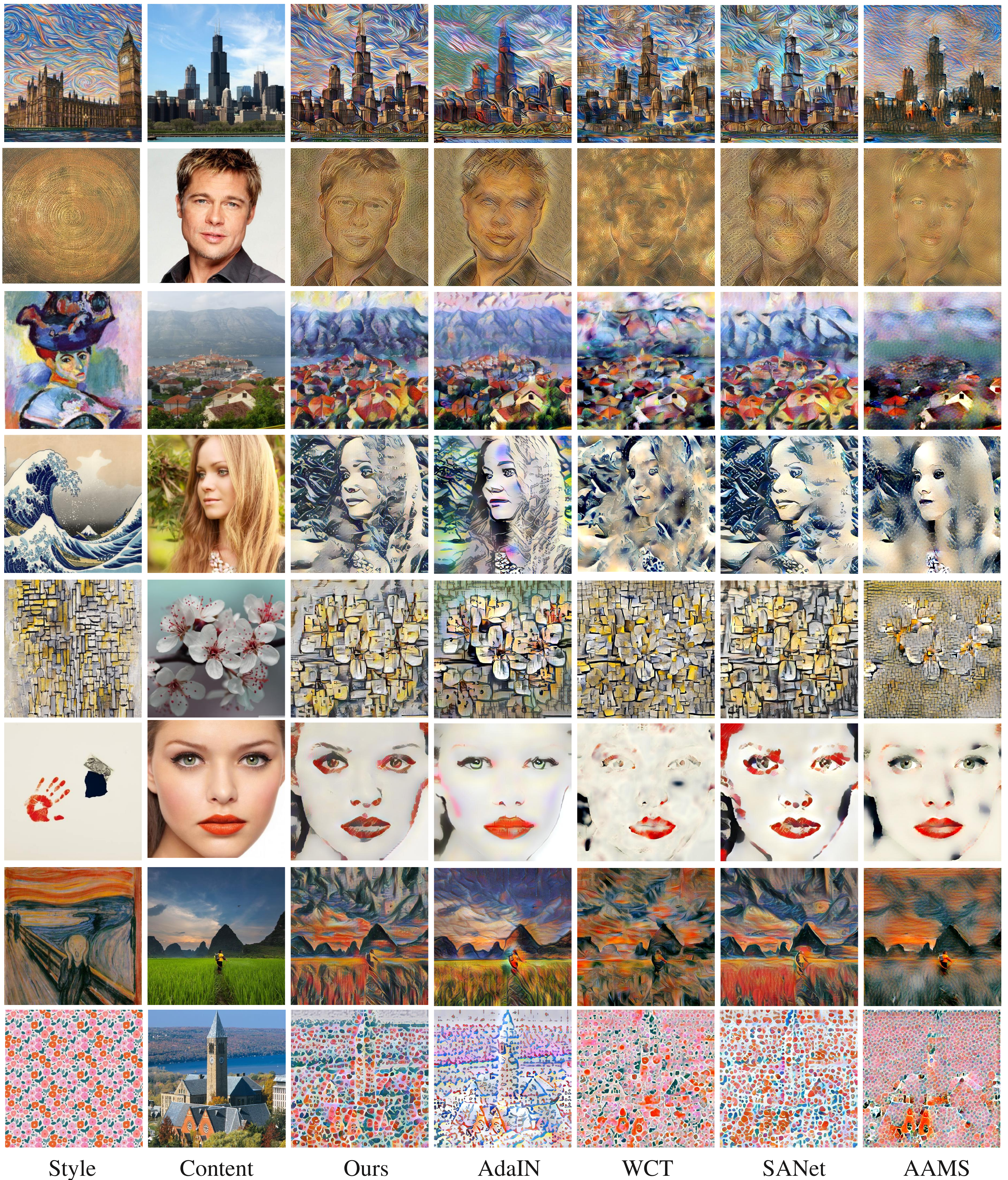}
\caption{Comparison of stylized results with SOTA methods. The first column shows style images, and the second column shows content images. The remaining columns are stylized results by our methods, AdaIN~\citep{Huang:2017:Arbitrary}, WCT~\citep{li:2017:universal}, SANet~\citep{park:2019:arbitrary}, and AAMS~\citep{yao:2019:attention}.}
\vspace{-4mm}
\label{fig:compare}
\end{figure*}
\begin{figure}[t]
\setlength{\abovecaptionskip}{2mm}
\setlength{\belowcaptionskip}{-2mm}
\centering
\includegraphics[width=\linewidth]{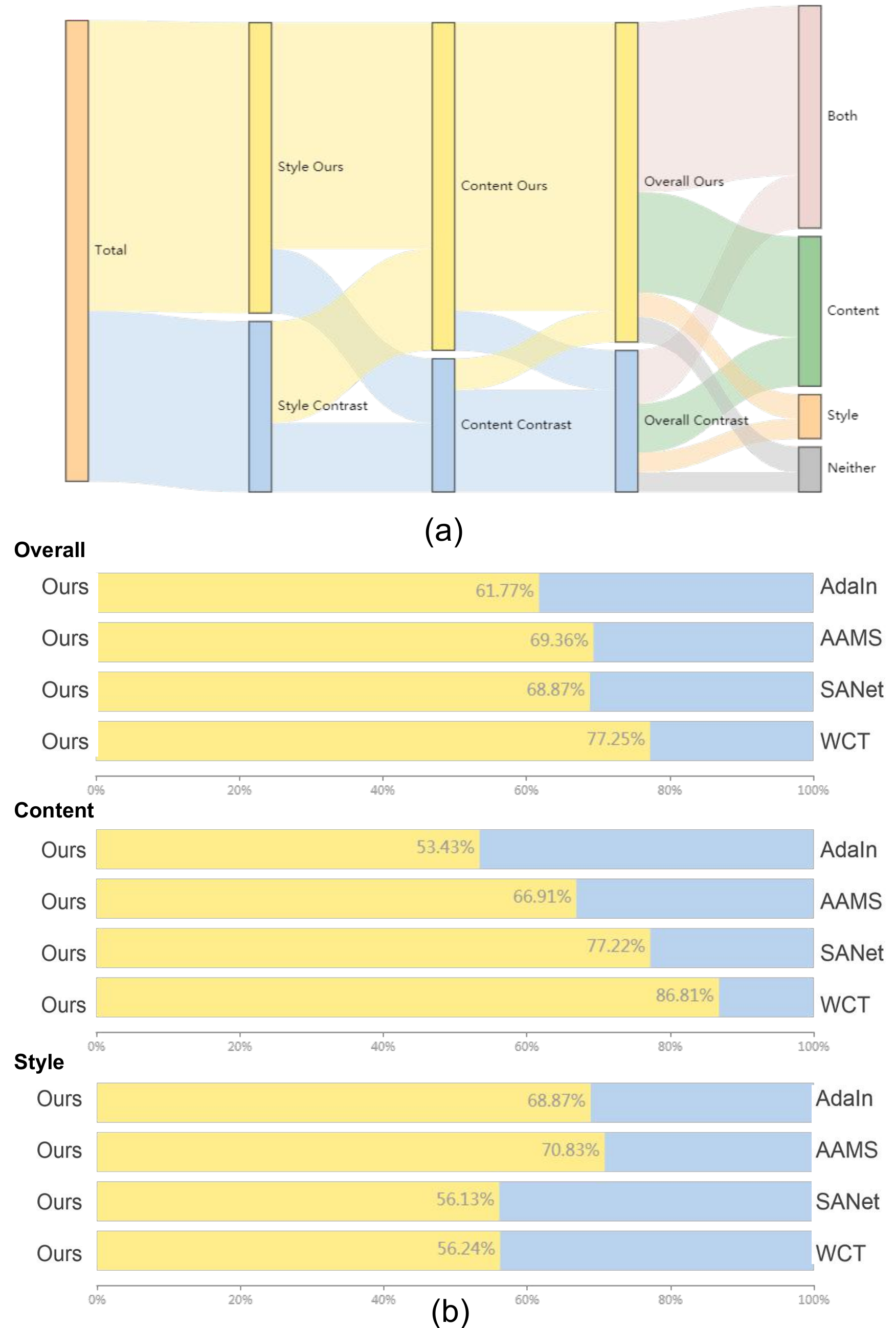}
\caption{User study results: (a) the Sankey diagram shows the overall results of users' preference between our results and the contrast results, (b) detailed results with each contrast method.}
\vspace{-2mm}
\label{fig:u2}
\end{figure}
\subsection{Implementation Details}

We use MS-COCO~\citep{lin:2014:coco} as content dataset and WikiArt~\citep{phillips:2011:wiki} as style dataset.
The style and content images are randomly cropped to $256 \times 256$ pixels in the training stage. 
All image sizes are supported in the testing stage. 
We use $conv1\_1$, $conv2\_1$, $conv3\_1$, and $conv4\_1$ layers in the encoder (pre-trained VGG19) to  extract image features. 
The features of $conv4\_1$ layer are fed to the multi-adaptation module to generate the features $f_{cs}$. 
Furthermore, we use layer $conv4\_1$ to calculate the content perceptual and disentanglement loss and $conv1\_1$, $conv2\_1$, $conv3\_1$, and $conv4\_1$ layers to calculate the style perceptual and disentanglement loss.
The convolution kernel sizes utilized in the multi-adaptation module are all set to $1 \times 1$.
The weights $\lambda_{c}$, $\lambda_{s}$, $\lambda_{id}$, $\lambda_{dis\_c}$, and $\lambda_{dis\_s}$ are set to $1$, $5$, $50$, $1$, and $1$, respectively.
\subsection{Comparison with Prior Work}
\label{sec:CPW}

\paragraph{Qualitative Evaluation}
We compare our method with four state-of-the-art works:  AdaIN~\citep{Huang:2017:Arbitrary}, WCT~\citep{li:2017:universal}, SANet~\citep{park:2019:arbitrary} and AAMS~\citep{yao:2019:attention}. 
Figure~\ref{fig:compare} shows the stylized results.. 
AdaIN~\citep{Huang:2017:Arbitrary} and WCT~\citep{li:2017:universal} adjust content images according to the second-order statistics of style images globally, but they ignore the local correlation between the content and style. Their stylized results have similar repeated textures in different image locations.
AdaIN~\citep{Huang:2017:Arbitrary} adjusts the mean and variance of the content image to adapt to the style image globally. 
 Inadequate textual patterns may be transferred in stylized images although the content structures are well preserved (\engordnumber{3} and \engordnumber{7}, rows in Figure~\ref{fig:compare}). 
Moreover, the results may even present different color distributions from style images (\engordnumber{1}, \engordnumber{4}, \engordnumber{5}, and \engordnumber{8} rows in Figure~\ref{fig:compare}). 
WCT~\citep{li:2017:universal} improves the style performance of AdaIN by adjusting the covariance of the content image through whitening and coloring transform operation. 
However, WCT would introduce content distortion (\engordnumber{2}, \engordnumber{4}, \engordnumber{6}, \engordnumber{7}, and \engordnumber{8} rows). 
SANet~\citep{park:2019:arbitrary} uses style attention to match the style features to the content features, which can generate attractive stylized results with distinct style texture. 
However, the content structures in stylized results are unclear without feature disentanglement (\engordnumber{2}, \engordnumber{4}, and \engordnumber{8} rows in Figure~\ref{fig:compare}). Moreover, the use of multi-layer features leads to repeated style patches in the results(eyes in the  \engordnumber{3} row in Figure~\ref{fig:compare}).
AAMS~\citep{yao:2019:attention} also adopts self-attention mechanism, but the use of self-attention is not effective enough. 
In the results, the main structures of the content images are clear, but the other structures are damaged and the style patterns in the generated image are not evident (\engordnumber{2}, \engordnumber{3}, \engordnumber{4}, \engordnumber{5}, and \engordnumber{8} rows in Figure~\ref{fig:compare}).

Disentangled content and style features in multi-adaptation network can well represent a domain-specific characteristic, which is different from the abovementioned methods. 
Therefore, the results generated by our method can further preserve the content and style information. 
Moreover, our method can generate good results, which have distinct content structures and rich style patterns, by adaptively adjusting the disentangled content and style features. 
The content images can be rendered by corresponding style patterns based on their semantic structures (\engordnumber{1} row in Figure~\ref{fig:compare}).

\begin{figure}
\setlength{\abovecaptionskip}{2mm}

\centering
\includegraphics[width=\linewidth]{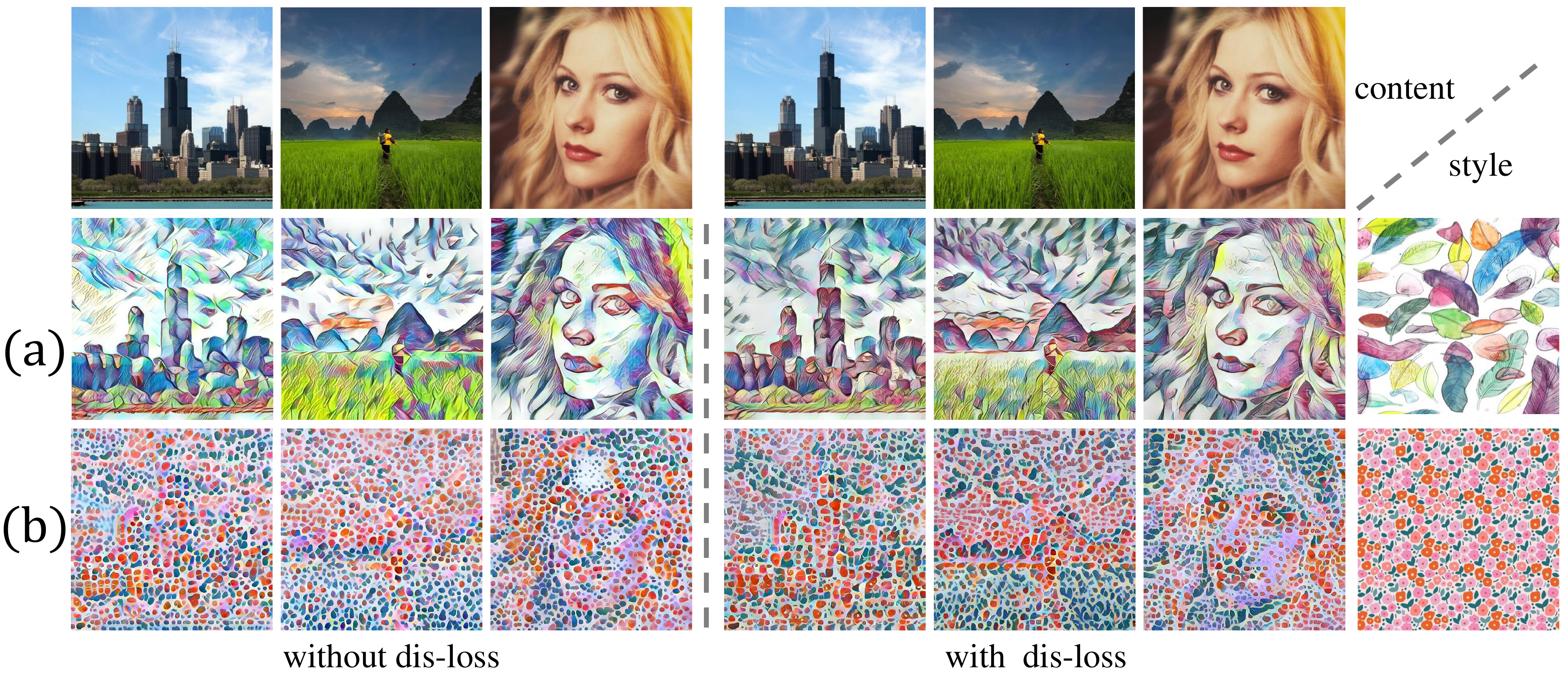}
\caption{Comparison of stylized results with/without disentanglement loss. (a) With style and content disentanglement, different content images have unified style patterns, which is the key component of style image(purple feathers). (b) With style and content disentanglement, the content structures are more visible. }
\vspace{-4mm}
\label{fig:dis}
\end{figure}
\paragraph{User Study}

We conduct user studies to compare the visual performance of ours and the aforementioned SOTA methods further.
We select $20$ style images and $15$ content images to generate $300$ results for each method. 
Initially, we show each content-style pair to the participants. 
Then we show two results (one is by our method and the other is randomly selected from one of the SOTA methods) to them.
We ask the participants four questions:
(1) which stylized result further preserves the content structures,
(2) which stylized result further transfers the style patterns,
(3) which stylized result has improved visual quality overall,
(4) when selecting the images in question (3), which factor is mainly considered: content, style, both, or neither?
We ask $30$ participants to do $50$ rounds of comparisons and obtain $1500$ votes for each question.
Figure~\ref{fig:u2} shows the statistical results.
Figure~\ref{fig:u2}(a) presents a Sankey diagram, which is used to demonstrate the direction of data flow. For example, among the users who select our method in style, most people also select our content, and a few select the content of the contrast methods. 
We can conclude from From Figure~\ref{fig:u2}(a) that regardless of the aspects of content, style, or overall effect, our method obtains the majority votes. 
The participants are more impressionable to content than style when selecting results with improved visual performance overall.

Subsequently, we compare our method with each comparison method in the aspect of content, style, and overall separately in Figure~\ref{fig:u2}(b).
The overall performance of our method is better than that of every comparison method.
Our results, compared with those of AdaIN~\citep{Huang:2017:Arbitrary}, has clear advantages in style and comparable content preservation ability.
Our results, compared with those of WCT~\citep{li:2017:universal}], have clear advantages in content and comparable style patterns. 
Our results, compared with those of AAMS~\citep{yao:2019:attention}, have clear advantages in content and style.

\begin{figure}
\setlength{\abovecaptionskip}{2mm}
\setlength{\belowcaptionskip}{-2mm}
\centering
\includegraphics[width=\linewidth]{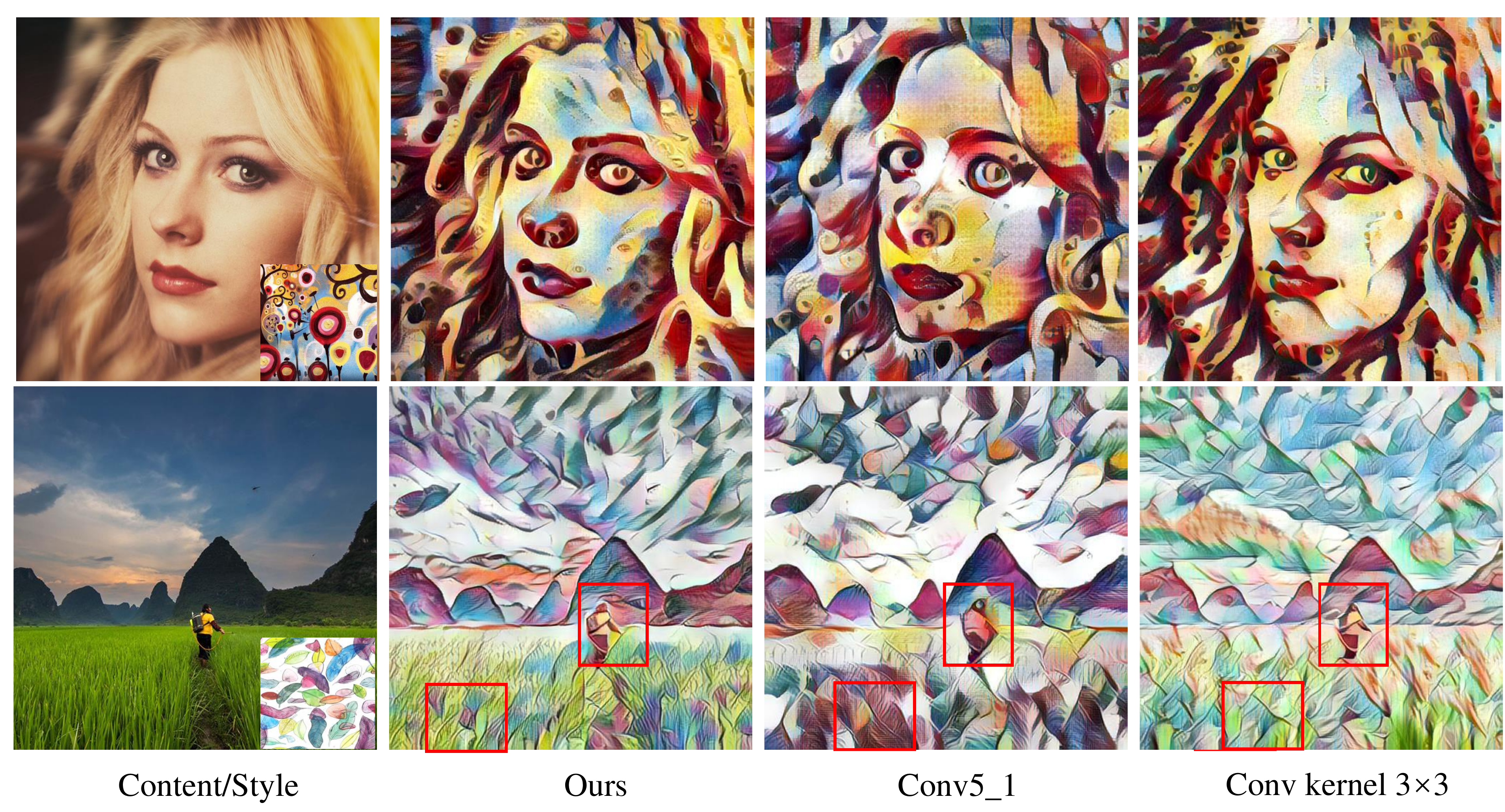}
\caption{Comparison of stylized results with different receptive field size. The meticulous structures of results in \engordnumber{3} and \engordnumber{4} columns are not well-preserved compared with our results (see details in red box).}
\vspace{-2mm}
\label{fig:a1}
\end{figure}
\begin{figure*}
\setlength{\abovecaptionskip}{2mm}

\centering
\includegraphics[width=\linewidth]{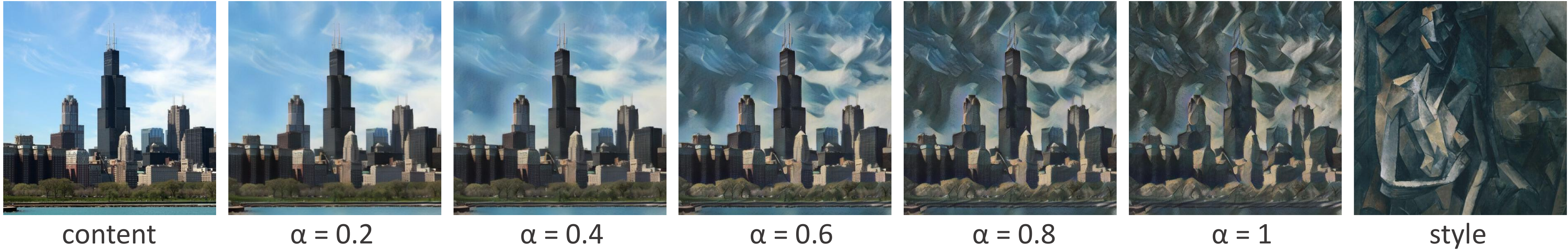}
\vspace{-4mm}
\caption{Trade-off between content and style.}
\vspace{-3mm}
\label{fig:weight}
\end{figure*}
\begin{figure*}
\setlength{\abovecaptionskip}{2mm}

\centering
\includegraphics[width=\linewidth]{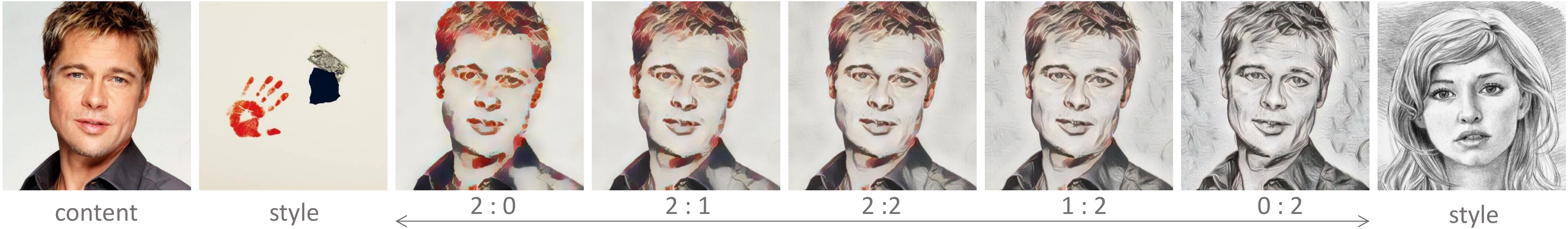}
\vspace{-3mm}
\caption{Style interpolation.}
\vspace{-4mm}
\label{fig:inter}
\end{figure*}

\paragraph{Quantitative Evaluation}
To quantify the style transferring and content preservation ability of our method, we first introduce an artist classification model to evaluate how well the stylized results are rendered by every artist's style. 
We select $5$ artists each with 1000 paintings and divide them into training and testing sets in the ratio of 8:2. Then we fine-tune the pretrained VGG19 model by using the training set.
We generate $1000$ stylized images for each method. 
We feed the stylized images to the artist classification model to calculate the accuracy. 
Second, we use a content classification model to quantify the content preservation effects of different methods. 
We randomly select $5$ classes from the ImageNet dataset.
Then we fine-tune the pretrained VGG19 model by using the training set of ImageNet.
We use the corresponding validation set (five classes, each class includes $50$ content images) to generate $1000$ stylized images for each method. 
We feed the stylized images to the content classification model to calculate the accuracy.

Table~\ref{tab:acc} shows the classification results..
The style and content classification accuracy values of our method are relatively high, which illustrates that our method can obtain a trade-off between content and style. 
Although the SANet achieves the highest style classification accuracy, the content classification accuracy is too low to obtain attractive results.  
The content classification accuracy of AdaIN is high, but the style classification accuracy is low. 
In general, the content/style classification results of each method are consistent with user study results. 
The small statistical difference is because participants may be influenced by the effect of generated results when they select improved content/style results.
\begin{table}

\centering
\caption{Classification accuracy.}
\vspace{-2mm}
\begin{tabular}{cccccc}
\toprule
&AdaIN& WCT& SANet& AAMS& Ours \\
\midrule 
style($\%$)&55.7& 61.5&\textbf{65.2}&57.9&62.9\\
\midrule 
content($\%$) &\textbf{42.0}& 22.8&29.1&33.0&34.6\\
\bottomrule
\end{tabular}
\vspace{-4mm}
\label{tab:acc}
\end{table}
\subsection{Ablation Study}

\paragraph{Verify the effect of disentanglement loss}
We compare the generated results with and without disentanglement loss to verify the effect of disentanglement loss.
As shown in Figure~\ref{fig:dis}, using disentanglement loss can generate results with the key style patterns of style image( purple feathers, Figure~\ref{fig:dis}(a)) or more visible content structures (Figure~\ref{fig:dis}(b)), compared with the stylized results without disentanglement loss. 
With disentanglement loss, the stylized results can preserve unified style patterns and salient content structure.

\paragraph{The Influence of Convolutional Receptive Field Size.}
The receptive field size of convolutional operation can influence the generated results when calculating the local similarity between disentangled content and style features in the CA module. 
There are two factors that can change the receptive field size.
First, the convolution kernel size is fixed to $3\times3$ in the encoder; the deeper the model is, the larger receptive field we can obtain.
Therefore, we use $conv5\_1$ by replacing $conv4\_1$ of the encoder layer to obtain a large receptive field.
Second, we feed the content and style features to two convolutional layers in the CA module and calculate their correlation. The convolutional kernel size used in this module is related to the receptive field, which influences the size of the region used to calculate the correlation. Thus, we change the convolutional kernel size from $1\times1$ to $3\times3$ to obtain a large receptive field.

As shown in Figure~\ref{fig:a1}, the stylized results using $conv5\_1$  or kernel size $3\times3$ are transferred to more local style patterns (e.g., circle patterns in the \engordnumber{1} row and feather patterns in the \engordnumber{2} row), and the content structures are highly distorted.
The results prove that a large receptive field pays further attention to the local structure and will spatially distort the global structure. 

\subsection{Applications}
\paragraph{Trade-off between content and style}
We can adjust the style pattern weights in the stylized results by changing $\alpha$ in the following function:
\begin{equation}
I_{cs} = \mathcal{D}( \alpha f_{cs} + (1- \alpha) f_c).
\end{equation}
When $\alpha = 0$, we obtain the original content image. When $\alpha = 1$, we obtain the fully  stylized image. We change $\alpha$ from $0$ to $1$.
Figure~\ref{fig:weight} shows the results.

\paragraph{Style Interpolation}
For a further flexible application, we can merge multiple style images into one generated result. 
Figure~\ref{fig:inter} presents the examples. We can also change the weights of different styles.

%% file: Conclusion.tex
\section{Conclusions and Future Work}
\label{sec:conclusion}

In this paper, we propose a multi-adaptation network to disentangle global content and style representation and adjust the style distribution to content distribution by considering the long-range local similarity between the disentangled content and style features.
Moreover, we propose a disentanglement loss to render the style features independent from the target content feature for constraining the separation of style and content features. 
Our method can achieve a trade-off between content structure preservation and style pattern rendering. 
Adequate experiments show that our network can consider the global content structure and local style patterns to generate fascinating results. 
We also analyze the effect of receptive field size in the CNNs on the generated results.

In future work, we aim to develop a style image selection method to recommend appropriate style images for a given content image based on the global semantic similarity of content and style for additional practical applications.